\pdfoutput=1

\documentclass[11pt]{article}

\usepackage[final]{acl}

\usepackage{times}
\usepackage{latexsym}

\usepackage[T1]{fontenc}

\usepackage[utf8]{inputenc}

\usepackage{microtype}

\usepackage{inconsolata}

\usepackage{graphicx}

\usepackage[hang,flushmargin]{footmisc}

\title{Conditioning LLMs with Emotion in Neural Machine Translation}

\author{Charles Brazier \and Jean-Luc Rouas \\
        Univ. Bordeaux, CNRS, Bordeaux INP, LaBRI, UMR 5800, F-33400 Talence, France \\
        \texttt{charles.brazier@u-bordeaux.fr} \hspace{0.5cm} \texttt{jean-luc.rouas@labri.fr}}

\begin{document}
\maketitle
\begin{abstract}
Large Language Models (LLMs) have shown remarkable performance in Natural Language Processing tasks, including Machine Translation (MT). In this work, we propose a novel MT pipeline that integrates emotion information extracted from a Speech Emotion Recognition (SER) model into LLMs to enhance translation quality. We first fine-tune five existing LLMs on the Libri-trans dataset and select the most performant model. Subsequently, we augment LLM prompts with different dimensional emotions and train the selected LLM under these different configurations. Our experiments reveal that integrating emotion information, especially arousal, into LLM prompts leads to notable improvements in translation quality.
\end{abstract}

\section{Introduction}
\label{sec:introduction}

Large Language Models (LLMs) are transformer-based \citep{vaswani2017attention} deep learning models designed to understand and generate natural language text by predicting the probability of the next token in a sequence. LLMs excel across various Natural Language Processing (NLP) tasks, such as information retrieval \citep{zhu2023large}, instruction following \citep{ouyang2022training}, or engaging in chatbot discussions \citep{chatgpt2022}.

Among NLP tasks, LLMs have shown great capacities in Machine Translation (MT) \citep{zhu2023multilingual}, the task of translating a text from one language to another. Previous research has enhanced LLM performance in MT through various strategies, including optimized prompting techniques \citep{zhang2023prompting}, in-context learning features \citep{brown2020language} to improve translation quality over time \citep{moslem2023adaptive, moslem2023fine}, and a two-stage fine-tuning method composed of a first fine-tuning on monolingual data to learn general linguistic knowledge followed by a second fine-tuning on parallel data \citep{xu2023paradigm} that establishes the current state-of-the-art method in MT.

Apart from LLMs, previous works in MT have demonstrated the possibility of controlling the translation by adding extra information to the model that is not explicitly specified in the source sentence to be translated, and that can influence the translation. Existing works in that direction focused on the control of politeness \citep{sennrich2019controlling}, gender \citep{vanmassenhove2018getting, gaido2023how}, or emotion \citep{brazier2024usefulness} of the translation and showed that this extra information helps improve translation quality.

In this work, we propose to improve translation performances of an LLM-based model by adding emotion as extra information in the prompt of the model to condition the translation. This work relies on the fact that words can be classified into emotion categories, leading to affective word lists \citep{pennebaker2001linguistic}. Thus, conditioning the translation with a specific emotion would use a suitable vocabulary in the translation. In \citet{brazier2024usefulness}, authors showed that adding arousal information, reflecting the level of stimulation (ranging from calm to excited), extracted from the voice and added at the start of each input text sentence, helps improve translation performances. In the following, we study the behavior of several LLMs for the task of MT when emotion dimensions are added to input prompts.

To address this problem, we first fine-tune several existing LLMs for the task of English-to-French text-to-text translation. Then, after selecting the best model as baseline for our experiments, we compute for each input sentence its emotional dimensions with the help of a state-of-the-art Speech Emotion Recognition (SER) model applied to audio recordings. Finally, we compare translation performance with and without the addition of each emotional dimension as extra information added to each input prompt. We show that emotion improves translation (BLEU and COMET), especially in the case of arousal.


\section{Related works}
\label{sec:Related_works}

In this work, we aim at combining an LLM-based MT model with emotion information to improve translation performances. In the following, we first describe a close work that performs this combination without the use of an LLM. Then, we list several existing LLMs that can be used as a baseline for our MT task.

\subsection{Machine Translation with Emotion}
\label{subsec:MT_Emotion}

To our knowledge, the only work that combines an MT model with emotion information is described in \citet{brazier2024usefulness}. In this study, the authors utilize a state-of-the-art Speech Emotion Recognition (SER) model \citep{wagner2023dawn} to automatically estimate dimensional emotion values, including arousal, dominance, and valence, for each audio recording associated with text sentence. These values are then transformed into unique emotion tokens, either positive or negative, which are added at the beginning of tokenized input text sentences. The authors report an increase in translation BLEU score, especially when adding arousal tokens at the start of input sentences.

The MT model used for their experiments is a transformer-based encoder-decoder architecture, comprising 6 layers for the encoder, 6 layers for the decoder, and 4 attention heads in each self-attention layer. The model is trained on the Libri-trans dataset \citep{kocabiyikoglu2018augmenting}, which includes triplets of English recordings, English texts, and French texts, totaling 235 hours of data (230h for train, 2h for dev, and 3.5h for test). The model performs English-to-French translation.

In this work, we propose to use the same translation pipeline, but instead of using a specific MT model, we replace it with a fine-tuned LLM. Since LLMs have more trainable parameters, we anticipate improved translation performances. However, our objective is to observe how LLMs behave when augmented with emotion information in the input prompt.

\subsection{LLM selection for MT}
\label{subsec:LLM4MT}

Recent advances in Large Language Modeling have significantly expanded the capabilities of LLMs across various tasks, such as reasoning, coding, or mathematics. Among the numerous existing LLMs \citep{chiang2024chatbot}, the best-performing models are GPT-4 \citep{openai2023gpt}, LLaMA 3 \citep{metaai2024llama}, Gemini 1.5 \citep{geminiteam2024gemini}, or Claude 3 \citep{anthropic2024claude}.

For the task of MT, we restrict our LLM selection to models that are open-source, promising (high rank in the LLM arena\footnote{\url{http://chat.lmsys.org/?leaderboard}}, or already fine-tuned to the MT task), and that only contain 7 billion (7B) of parameters. We select 5 different models that are described in the following.

The first selected LLM is \textit{Mistral-7B-v0.1}\footnote{\url{http://huggingface.co/mistralai/Mistral-7B-v0.1}}, an open-source model \citep{jiang2023mistral} which ranks among the best 7B-parameter models.

As the second model, we select \textit{Mistral-7B-Instruct-v0.2}\footnote{\url{http://huggingface.co/mistralai/Mistral-7B-Instruct-v0.2}}. The model is similar to the previous model but has been fine-tuned to follow instructions.

Our third selected model is \textit{TowerBase-7B-v0.1}\footnote{\url{http://huggingface.co/Unbabel/TowerBase-7B-v0.1}}. This model \citep{alves2024tower} is based on LLaMA 2 \citep{metaai2023llama} and its training has been continued on multilingual data (including English and French monolingual data, as well as bilingual data).

Similarly to Mistral, we select \textit{TowerInstruct-7B-v0.2}\footnote{\url{http://huggingface.co/Unbabel/TowerInstruct-7B-v0.2}} as our fourth model. This model is a variant of the previous one that has been fine-tuned to follow instructions including translations.

Finally, as our fifth model, we select the SOTA MT model \textit{ALMA-7B-R}\footnote{\url{http://huggingface.co/haoranxu/ALMA-7B-R}}, which is based on LLaMA 2 \citep{metaai2023llama}, and fine-tuned on monolingual and parallel data. However, the data used for fine-tuning does not include French.

\section{Experiments and results}
\label{sec:Experiments}

In this section, we describe our experiments for the task of English-to-French text-to-text translation. We conduct two successive experiments. Firstly, we fine-tune five existing LLMs on the Libri-trans dataset \cite{kocabiyikoglu2018augmenting} and consider the best model as a foundation for our second experiment. Secondly, we fine-tune the selected LLM on the same task but under different configurations. Henceforth, prompts used for translation include each emotion dimension that is automatically estimated from the SER model.

\subsection{Fine-tuning LLMs on Libri-trans}
\label{subsec:LLM_libritrans}

To perform MT with LLMs, the task needs to be converted into a language modeling problem with the use of prompts. In this work, we perform zero-shot prompting and follow two different templates. The first template will be applied to \textit{Mistral-7B-v0.1} and \textit{TowerBase-7B-v0.1}:
\begin{equation}
    \texttt{\small English: <src txt> \textbackslash n French: <tgt txt>}
\end{equation}
where \texttt{<src txt>} and \texttt{<tgt txt>} refer to the English source sentence and the French target sentence respectively.

The second template will be applied to models that follow instructions, namely \textit{Mistral-7B-Instruct-v0.2}, \textit{TowerInstruct-7B-v0.2}, and \textit{ALMA-7B-R}:
\begin{equation}
    \texttt{\tiny [INST] Translate from English to French: <src txt> [/INST] \textbackslash n <tgt txt>}
\end{equation}

To fine-tune LLMs, we employ QLoRA \citep{hu2022lora, dettmers2023qlora}, a Parameter Efficient Fine-Tuning method \cite{mangrulkar2022peft} that allows training with significantly fewer parameters. Additionally, we apply a 4-bit quantization to reduce memory usage while maintaining 16-bit precision during computation.

We provide two distinct metrics to evaluate our MT models. The first metric is the BLEU score computed using sacrebleu \citep{post2018call}. It reflects the degree of lexical matches (number of common n-grams) between the proposed translation and its corresponding reference. The second metric is the COMET score \footnote{\url{https://huggingface.co/Unbabel/wmt22-comet-da}} \citep{rei2022comet}. It is computed from a trained model and reflects translation quality between translation, reference, and also the source sentence. According to the metric ranking presented in \citet{freitag2022results}, we rely more on the COMET score than on the BLEU score.

Table~\ref{tab:LLM_libritrans} showcases the results of our first experiment. In this table, we report BLEU and COMET scores of the five selected LLMs on both the dev and test sets of the Libri-trans dataset.

\begin{table}
  \centering
  \begin{tabular}{lcccc}
    \hline
    \textbf{Model} & \multicolumn{2}{c}{\textbf{BLEU}} & \multicolumn{2}{c}{\textbf{COMET}} \\
    & \textbf{dev} & \textbf{test} & \textbf{dev} & \textbf{test} \\
    \hline
    Mistral         & 16.4 & 16.7 & 73.2 & 72.5\\
    MistralInstruct & 16.0 & 17.9 & 72.1 & 71.9\\
    TowerBase       & \textbf{24.0} & \textbf{20.6} & \textbf{73.8} & \textbf{72.9}\\
    TowerInstruct   & 6.4 & 6.1 & 35.5 & 35.5\\
    ALMA            & 7.1 & 7.5 & 52.1 & 52.8\\
    \hline
  \end{tabular}
  \caption{BLEU and COMET scores of our five selected LLMs on dev and test sets of Libri-trans.}
  \label{tab:LLM_libritrans}
\end{table}

The table highlights three models, \textit{Mistral-7B-v0.1}, \textit{Mistral-7B-Instruct-v0.2}, and \textit{TowerBase-7B-v0.1}, that attain high BLEU and COMET scores. They obtain COMET scores ranging from 72.1 to 73.8 on the dev set and from 71.9 to 72.9 on the test set. Additionally, their BLEU scores ranged from 16.0 to 24.0 on the dev set and from 16.7 to 20.6 on the test set. While COMET scores are not meant to be interpretable (but enable the comparison between models), BLEU scores indicate, on average, a translation that is more or less clear with numerous grammatical errors. These low BLEU scores are comparable to performances of previous works on this dataset \citep{zhao2021neurst, brazier2024usefulness} and are mainly caused by the nature of the data (audiobooks with literary vocabulary).

Also, it is worth noting that two models, \textit{TowerInstruct-7B-v0.2} and \textit{ALMA-7B-R}, exhibit poor performances in MT when fine-tuned on Libri-trans. In the case of \textit{ALMA-7B-R}, this can be explained by the fact that French is not among the languages included in the data used to pre-train the model. Thus, the model fails at predicting French text.

As additional training information, all LLMs have obtained their optimal state in a maximum of 5 epochs. This represents a training time of 3 hours on a GPU NVIDIA A100 for each model. This fast fine-tuning time is due to QLoRA and 4-bit quantization strategies.

To summarize, the best machine translation performances were achieved with the \textit{TowerBase-7B-v0.1}. This LLM serves as a baseline and foundation model for the following experiment.

\subsection{Fine-tuning LLMs with Emotion}
\label{subsec:LLM_emotion}

The second experiment aims at observing the behavior of our LLM-based \textit{TowerBase-7B-v0.1} model on the task of English-to-French Machine Translation when emotion information is added to the prompt before translation.

As a first step, we estimate the emotion of each English recording present in the Libri-trans dataset. Following the same methodology as \citet{brazier2024usefulness}, we compute dimensional emotion values for arousal, dominance, and valence with the help of a trained SER model \citep{wagner2023dawn}. Emotion values range between 0 and 1 and are correctly balanced (medians between 0.4 and 0.6, see \citet{brazier2024usefulness}).

As a second step, we create specific prompts that include the emotion information in the text. For this purpose, we propose 3 different templates. The first template adds emotion information before the source sentence:
\begin{equation} \label{eq:1}
    \texttt{\scriptsize English <status> <emotion>: <src txt> \textbackslash n French: <tgt txt>}
\end{equation}
where \texttt{status} is replaced by either \textit{with} or \textit{without} if the emotion value is higher or lower than 0.5 respectively, \texttt{emotion} is replaced by either \textit{arousal}, \textit{dominance}, or \textit{valence}, \texttt{src txt} represents the English source sentence, and \texttt{tgt txt} represents the French target translation.

The second template adds emotion information before the target sentence:
\begin{equation} \label{eq:2}
    \texttt{\scriptsize English: <src txt> \textbackslash n French <status> <emotion>: <tgt txt>}
\end{equation}

The third template is inspired from \citet{brazier2024usefulness}, where emotion information is added as a discrete token at the start of the source sentence:
\begin{equation} \label{eq:3}
    \texttt{\scriptsize English: [<emotion> <polarity>] <src txt> \textbackslash n French: <tgt txt>}
\end{equation}
where \texttt{polarity} is replaced by either \textit{positive} or \textit{negative} if the emotion value is higher or lower than 0.5 respectively.

In this experiment, the \textit{TowerBase-7B-v0.1} model is retrained from its initial state and not from the training checkpoint obtained after the previous experiment. In the following, all models obtain their best performances in less than 5 training epochs.

Table~\ref{tab:LLM_emotion} showcases the results of our second experiment. It reports BLEU and COMET scores of the selected \textit{TowerBase-7B-v0.1} model on the dev and test sets of the Libri-trans dataset under different configurations. The first line mentions the score of the LLM obtained in the previous experiment and serves as a baseline for the second experiment. The other lines correspond to the model trained with different emotions (arousal, dominance, or valence), and with different prompts (the numbers \ref{eq:1}, \ref{eq:2}, and \ref{eq:3} refer to their equation number).

\begin{table}
  \centering
  \begin{tabular}{lcccc}
    \hline
    \textbf{Model} & \multicolumn{2}{c}{\textbf{BLEU}} & \multicolumn{2}{c}{\textbf{COMET}} \\
    & \textbf{dev} & \textbf{test} & \textbf{dev} & \textbf{test} \\
    \hline
    TowerBase         & 24.0 & 20.6 & 73.8 & 72.9\\
    ~~~+arousal\ref{eq:1}      & 22.1 & 21.8 & \textbf{74.9} & \textbf{74.3}\\
    ~~~+arousal\ref{eq:2}      & \textbf{25.6} & \textbf{24.1} & 74.8 & 73.9\\
    ~~~+arousal\ref{eq:3}      & 19.3 & 19.2 & 74.2 & 73.4\\
    ~~~+dominance\ref{eq:1}    & 19.9 & 19.4 & 74.4 & 73.5\\
    ~~~+dominance\ref{eq:2}    & 18.9 & 20.9 & \textbf{74.9} & 74.0\\
    ~~~+dominance\ref{eq:3}    & 16.5 & 20.1 & 73.4 & 73.0\\
    ~~~+valence\ref{eq:1}      & 21.5 & 18.9 & 74.1 & 73.5\\
    ~~~+valence\ref{eq:2}      & 18.3 & 21.2 & 74.6 & 73.9\\
    ~~~+valence\ref{eq:3}      & 17.2 & 16.0 & 74.5 & 73.6\\
    \hline
  \end{tabular}
  \caption{BLEU and COMET scores of the TowerBase model on dev and test sets of Libri-trans. First line: baseline score. Other lines: score when trained with emotion in the prompt.}
  \label{tab:LLM_emotion}
\end{table}

We first remark that, except in the case of \textit{dominance\ref{eq:3}}, all COMET scores improved, compared to their baseline. This reflects a better translation quality when adding emotion information to the prompts. The best COMET scores are obtained when arousal information is added to the prompt using Equation~\ref{eq:1}. In this configuration, COMET scores are increased by +1.1 and +1.4 for the dev and test sets of Libri-trans respectively.

Secondly, we observe that BLEU scores show improvements only for specific models. The best BLEU scores are obtained when arousal information is added to the prompt using Equation~\ref{eq:2}. In this configuration, BLEU scores increase by +1.6 and +3.5 for the dev and test sets of Libri-trans respectively. However, due to the low ranking of BLEU \citep{freitag2022results}, we do not conduct further analysis based on this metric.

In summary, incorporating emotion information into the translation process appears to enhance translation quality. The highest scores are achieved when utilizing the arousal dimension with Equation~\ref{eq:1} or \ref{eq:2}. This finding aligns with the results reported in \citet{brazier2024usefulness}.

\section{Conclusion}
\label{sec:Conclusion}

We proposed a new MT pipeline that combines an LLM-based model and emotion information extracted from a SER model to improve translation performances. We obtain the best performances when the arousal value is added to the LLM prompt.

As future work, we will apply our method to other multilingual datasets including Must-C \citep{di2019must}. Unlike the Libri-trans dataset, which consists of literary text read by speakers, Must-C encompasses various speech types, such as TED talks, which can offer more emotional variability and therefore further enhance translation performance. We also plan to extend our method to the speech-to-text task, also known as Speech translation.

\section{Acknowledgements}
\label{sec:Acknowledgements}

The research presented in this paper is conducted as part of the project FVLLMONTI, which has received funding from the European Union’s Horizon 2020 Research and Innovation action under grant agreement No 101016776.

\bibliography{custom}

\end{document}